\documentclass{article}
\usepackage{microtype}
\usepackage{multicol}
\usepackage{adjustbox}
\usepackage{multirow}
\usepackage{graphicx}
\usepackage{subfigure}
\usepackage{booktabs}
\usepackage{hyperref}
\usepackage{amsfonts}
\usepackage{amsmath}
\usepackage{xcolor}
\usepackage{tikz}
\usepackage{lipsum}
\usepackage{cleveref}
\usepackage{enumitem}
\usepackage[round]{natbib}
\usepackage{arxiv}

\hypersetup{
    colorlinks,
    linkcolor={blue!50!black},
    citecolor={blue!50!black},
    urlcolor={blue!50!black}
}

\newcommand{\benchmark}{\textsc{UIMNET}}

\title{What classifiers know what they don't?}
\author{Mohamed Ishmael Belghazi and David Lopez-Paz\\
\texttt{\{ishmaelb, dlp\}@fb.com}\\
Facebook AI Research, Paris, France}

\begin{document}

\maketitle

\begin{abstract}
Being uncertain when facing the unknown is key to intelligent decision making.
However, machine learning algorithms lack reliable estimates about their predictive uncertainty.
This leads to wrong and overly-confident decisions when encountering classes unseen during training.
Despite the importance of equipping classifiers with uncertainty estimates ready for the real world, prior work has focused on small datasets and little or no class discrepancy between training and testing data.
To close this gap, we introduce \benchmark{}: a realistic, ImageNet-scale test-bed to evaluate predictive uncertainty estimates for deep image classifiers.
Our benchmark provides implementations of eight state-of-the-art algorithms, six uncertainty measures, four in-domain metrics, three out-domain metrics, and a fully automated pipeline to train, calibrate, ensemble, select, and evaluate models.
Our test-bed is open-source and all of our results are reproducible from a fixed commit in our repository.
Adding new datasets, algorithms, measures, or metrics is a matter of a few lines of code---in so hoping that \benchmark{} becomes a stepping stone towards realistic, rigorous, and reproducible research in uncertainty estimation.
Our results show that ensembles of ERM classifiers as well as single MIMO classifiers are the two best alternatives currently available to measure uncertainty about both in-domain and out-domain classes.
\end{abstract}

\section{Introduction}
\begin{flushright}
  \emph{I don't think I've ever seen anything quite like this before}\\
  ---HAL 9000 in \textit{2001: A Space Odyssey}
\end{flushright}

Deep image classifiers exceed at discriminating the set of \emph{in-domain} classes observed during training.
However, when confronting test examples from unseen \emph{out-domain} classes, these classifiers can only predict in terms of their known in-domain categories, leading to wrong and overly-confident decisions \citep{hein2019relu, ulmer2020know}.
In short, machine learning systems are unaware of their own limits, since ``they do not know what they do not know''.
Because of this reason, out-domain data cannot be safely identified and treated accordingly.
Thus, it is reasonable to fear that, when deployed in-the-wild, the behavior of these classifiers becomes unpredictable and their performance crumbles by leaps and bounds \citep{ovadia2019can}.

The inability of machine learning systems to estimate their uncertainty and abstaining to classify out-domain classes is roadblock towards their implementation in critical applications.
These include self-driving \citep{michelmore2018evaluating}, medicine \citep{begoli2019need}, and the analysis of satellite imagery \citep{wadoux2019using}.
Good uncertainty estimates are also a key ingredient in anomaly detection~\citep{chalapathy2019deep}, active learning~\citep{settles2009active}, safe reinforcement learning \citep{henaff2019model}, defending against adversarial examples \citep{goodfellow2014explaining}, and model interpretability \citep{alvarez2017causal}.
For an extensive literature review on uncertainty estimation and its applications, we refer the curious reader to the survey of~\citet{abdar2020review} and the one of \citet{ruff2021unifying}.
Despite a research effort spanning multiple decades, machine learning systems still lack trustworthy estimates of their predictive uncertainty.
In our view, one hindrance to this research program is the absence of realistic benchmarking and evaluation protocols.
More specifically, prior attempts are limited in two fundamental ways.
First, these experiment on small datasets such as SVHN and CIFAR-10 \citep{van2021improving}.
Second, these do not provide a challenging set of out-domain data.
Instead, they construct out-domain classes by using a second dataset (e.g., using MNIST in-domain versus FashionMNIST out-domain, cf.~\citet{van2020uncertainty}) or by perturbing the in-domain classes using handcrafted transformations (such as Gaussian noise or blur, see ImageNet-C \citet{hendrycks2019benchmarking}).
Both approaches result in simplistic benchmarking, and little is learned about uncertainty estimation for the real world.
The purpose of this work is to introduce an end-to-end benchmark and evaluation protocol bridging this disconnect.
At the time of writing, \benchmark{} is the most exhaustive benchmark for uncertainty estimation in the literature.

\begin{table*}[t!]
\begin{center}
\begin{tabular}{lllllll}
    \toprule
    \textbf{Datasets} & \textbf{Algorithms}   & \textbf{Uncertainty}      & \textbf{In-Domain}  & \textbf{Out-Domain} & \textbf{Ablations} \\
                      &                       & \textbf{measures}         & \textbf{metrics}    & \textbf{metrics}    & \\
    \midrule                                                         
    ImageNot          & ERM                   & Largest                   & ACC@1               & AUC                 &Calibration (y/n)\\
                      & Mixup                 & Gap                       & ACC@5               & InAsIn              &Spectral norm (y/n)\\
                      & Soft-labeler          & Entropy                   & ECE                 & InAsOut             &Model size (RN18 / RN50)\\
                      & RBF                   & Jacobian                  & NLL                 & OutAsIn             &\\
                      &  RND                  & GMM                       &                     & OutAsOut            &\\
                      &  OCs                  & Native                    &                     &                     &\\
                      &  MC-Dropout           &                           &                     &                     &\\
                      &  MIMO                 &                           &                     &                     &\\
                      &  (+ Ensembles)        &                           &                     &                     &\\
    \bottomrule
\end{tabular}
\label{table:summary}
\end{center}
\caption{The \benchmark{} test-bed suite for uncertainty estimation.}
\end{table*}

\paragraph{Formal setup}
Following conventional supervised learning, we learn a classifier $f$ using \emph{in-domain} data from the distribution $P_\text{in}(X, Y)$.
After training, we endow the classifier with a real-valued uncertainty measure $u(f, x^\dagger)$.
Given a test example $(x^\dagger, y^\dagger) \sim P$ with unobserved label $y^\dagger$, we declare $x^\dagger$ \emph{in-domain} (hypothesizing $P = P_\text{in}$) if $u(f, x^\dagger)$ is small, whereas we declare $x^\dagger$ \emph{out-domain} (hypothesizing $P \neq P_\text{in}$) if $u(f, x^\dagger)$ is large.
Using these tools, our goal is to abstain from classifying out-domain test examples, and to classify with calibrated probabilities in-domain test examples.
The sequel assumes that the difference between in- and out-domain resides in that the two groups of data concern disjoint classes.

\paragraph{Contributions}
We introduce \benchmark{}, a test-bed for large-scale, realistic evaluation of uncertainty estimates in deep image classifiers.
We outline the components of the test-bed below, also summarized in Table~\ref{table:summary}.
{
\setlist[itemize]{leftmargin=16mm}

\begin{itemize}
  \item[(Sec. \ref{sec:datasets})] We construct ImageNot, a perceptual partition of ImageNet into \emph{in-domain} and \emph{out-domain} classes.
  Unlike prior work focused on small datasets like SVHN and CIFAR-10, ImageNot provides a benchmark for uncertainty estimators at a much larger scale.
  Moreover, both in-domain and out-domain categories in ImageNot originate from the original ImageNet dataset.
  This provides realistic out-domain data, as opposed to prior work relying on a second dataset (e.g., MNIST as in-domain versus SVHN as out-domain), or handcrafted perturbations of in-domain classes (Gaussian noise or blur as out-domain).

  \item[(Sec. \ref{sec:algorithms})] We re-implement eight state-of-the-art algorithms from scratch, listed in Table~\ref{table:summary}.
  This allows a fair comparison under the exact same experimental conditions (training/validation splits, hyper-parameter search, neural network architectures and random initializations).
  Furthermore, we also study ensembles of multiple training instances for each algorithm.
  \item[(Sec. \ref{sec:measures})] Each algorithm can be endowed with one out of six possible uncertainty measures, allowing an exhaustive study of what algorithms play well with what measures.
  Listed in Table~\ref{table:summary}, these are the largest softmax score, the gap between the two largest softmax scores, the softmax entropy, the norm of the Jacobian, a per-class Gaussian density model, and (for those available) an algorithm-specific measure.
  \item[(Sec. \ref{sec:metrics})] For each classifier-measure pair, we study four in-domain metrics (top-1 and top-5 classification accuracy, log-likelihood, expected calibration error) and three out-domain metrics (the AUC at classifying in-domain versus out-domain samples using the selected uncertainty measure, as well as the confusion matrix at a fixed uncertainty threshold computed over an in-domain validation set).
  \item[(Sec. \ref{sec:ablations})] We repeat our entire pipeline to accommodate three popular ablations to understand the impact of model calibration by temperature scaling, model size, and the use of spectral normalization. 
  \item[(Sec. \ref{sec:experiments})] \benchmark{} is entirely hands-off, since the pipeline from zero to \LaTeX{} tables is fully automated: this includes hyper-parameter search, model calibration, model ensembling, and the production of all the tables included in our experimental results.
  \item[(Sec. \ref{sec:conclusion})] Our experimental results illustate that no classifier but those trained by using Multiple-Input and Multiple-Output (MIMO) systematically dominates calibrated classifiers trained by empirical risk minimization. Interestingly, a single MIMO classifier \emph{if given enough capacity} matches ensembles of ERMs in terms of AUC score and identification of in-domain instances and even edges them when identifying out-domaim instances. Additionally, our result show that the use of spectral normalization does not significantly improve out of domain detection metrics while the effect of calibration remains mild at best.
\end{itemize}
}  
\benchmark{} is open sourced at \url{https://github.com/facebookresearch/uimnet}.
All of the tables presented in this paper are reproducible by running the main script in the repository at commit \texttt{0xSOON}.
  
\section{Constructing the ImageNot benchmark}
\label{sec:datasets}

The ImageNet dataset \citep{russakovsky2015imagenet}, a gold standard to conduct research in supervised learning for computer vision, pertains the classification of images into $1000$ different classes.
Here we will use the ImageNet dataset to derive ImageNot, a large-scale and realistic benchmark for uncertainty estimation.
In a nutshell, we will divide the $1000$ classes of the original ImageNet dataset into \emph{in-domain} classes (used to train and evaluate algorithms in-distribution) and \emph{out-domain} classes (used to evaluate algorithms out-of-distribution).

To partition ImageNet into in-domain and out-domain, we featurize the entire dataset to understand the perceptual similarity between classes.
To this end, we use a pre-trained ResNet-18~\citep{he2016deep} to compute the average last-layer representation for each of the classes.
Next, we use agglomerative hierarchical clustering \citet{ward1963hierarchical} to construct a tree describing the perceptual similarities between the $1000$ average feature vectors.
Such perceptual tree has $1000$ leafs, each of them being a cluster containing one of the classes.
During each step of the iterative agglomerative clustering algorithm the two closest clusters are merged, where the distance between two clusters is computed using the criterion of \citet{ward1963hierarchical}.
The algorithm halts when there are only two clusters left to merge, forming the root node of the tree.

At this point, we declare the $266$ classes to the left of the root as \emph{in-domain}, and the first $266$ classes to the right of the root as \emph{out-domain}.
In the sequel, we call ``training set'' and ``validation set'' to a 90/10 random split from the original ImageNet ``train'' set. We call ``testing set'' to the original ImageNet ``val'' split.
The exact in-/out-domain class partition as well as the considered train/validation splits are specified in Appendix~\ref{app:partitions}.

While inspired by the BREEDS dataset \citep{santurkar2020breeds}, our benchmark ImageNot is conceived to tackle a different problem.
The aim of the \textsc{BREEDS} dataset is to classify ImageNet into a {small} number of super-classes, each of them containing a number of perceptually-similar sub-classes.
The \textsc{BREEDS} training and testing distributions differ on the sub-class proportions contributing to their super-classes.
Since the \textsc{BREEDS} task is to classify super-classes, the set of labels remains constant from training to testing conditions.
This is in contrast to ImageNot, where the algorithm observes only in-domain classes during training, but both in-domain and out-domain classes during evaluation.
While \textsc{BREEDS} studies the important problem of domain generalization~\citep{gulrajani2020search}, where there is always a right prediction to make within the in-domain classes during evaluation, here we focus on measuring uncertainty and abstaining from out-domain classes unseen during training.

Also related, our benchmark ImageNot is similar to the ImageNet-O dataset of \citep{hendrycks2021natural}. However, their out-domain classes are obtained from heterogeneous sources outsider of ImageNet, which have distinct statistics that can be easily picked up by the classifier.
In contrast, the starting point for both in-domain and out-domain classes of our ImageNot is the same (the original ImageNet dataset) and thus should maximally overlap in terms of image statistics, leading to a more challenging, realistic benchmark.

\section{Algorithms}
\label{sec:algorithms}

We benchmark eight
supervised learning algorithms that are commonly applied to tasks involving uncertainty estimation.
Each algorithm consumes one in-domain training set of image-label pairs $\{(x_i, y_i)\}_{i=1}^n$ and returns a \emph{predictor}\footnote{In the sequel, we denote by \emph{classifier} the last few layers in the \emph{predictor} that follow after the \emph{featurizer}.} $f(x) = w(\phi(x))$, composed by a \emph{featurizer} $\phi : \mathbb{R}^{3 \times 224 \times 224} \to \mathbb{R}^k$ and a \emph{classifier} $w : \mathbb{R}^k \to \mathbb{R}^C$.
We consider predictors implemented using deep convolutional neural networks \citep{lecun2015deep}.
Given an input image $x^\dagger$, all predictors return a softmax vector $f(x^\dagger) = (f(x^\dagger)_c)_{c=1}^C$ over $C$ classes.
The considered algorithms are:
\begin{itemize}
  \item \textbf{Empirical Risk Minimization}, or ERM \citep{vapnik1992principles}, or vanilla training.
  \item \textbf{Mixup} \citep{zhang2017mixup} chooses a predictor minimizing the empirical risk on \emph{mixed} examples ${(\tilde{x}, \tilde{y})}$, built as: 
    \begin{align*}
        \lambda   &\sim \text{Beta}(\alpha, \alpha),\\
        \tilde{x} &\sim \lambda \cdot x_i + (1 - \lambda) \cdot x_j,\\
        \tilde{y} &\sim \lambda \cdot y_i + (1 - \lambda) \cdot y_j,
    \end{align*}
  where $\alpha$ is a mixing parameter, and $((x_i, y_i), (x_j, y_j))$ is a random pair of training examples.
  Mixup has been shown to improve both generalization performance \citep{zhang2017mixup} and calibration  error \citep{thulasidasan2019mixup}.
  \item \textbf{Random Network Distillation}, or RND \citep{burda2018exploration}, finds an ERM predictor $f(x) = w(\phi(x))$, but also trains an auxiliary classifier $w_\text{student}$ to minimize $$\|w_\text{student}(\phi(x)) - w_\text{teacher}(\phi(x))\|^2_2,$$ where $w_\text{teacher}$ is a fixed classifier with random weights.
  RND has shown good performance as a tool for exploration in reinforcement learning.
  \item \textbf{Orthogonal Certificates}, or OC \citep{tagasovska2018single}, is analogous to RND for $w_\text{teacher}(\phi(x)) = \vec{0}_k$ for all $x$.
  That is, the goal of $w_\text{student}$ is to map all the in-domain training examples to zero in $k$ different ways (or \emph{certificates}).
  To ensure diverse and non-trivial certificates, we regularize each weight matrix $W$ of $w_\text{student}$ to be orthogonal by adding a regularization term $\|W^\top W - I\|_2^2$.
  OCs have shown good performance at the task of estimating uncertainty across a variety of classification tasks.
  \item \textbf{MC-Dropout} \citep{gal2016dropout} uses ERM over a family of predictors with one or more dropout layers \citep{srivastava2014dropout}.
  These stochastic dropout layers remain active at test time, allowing the predictor to produce multiple softmax vectors $\{f(x^\dagger, \text{dropout}_t)\}_{t=1}^T$ for each test example $x^\dagger$.
  Here, $\text{dropout}_t$ is a random dropout mask sampled anew.
  MCDropout is one of the most popular baselines to estimate uncertainty. 
  \item \textbf{MIMO} \citep{havasi2021training} is a variant of ERM over predictors accepting $T$ images and producing $T$ softmax vectors.
  For example, MIMO with $T=3$ is trained to predict jointly the label vector $(y_i, y_j, y_k)$ using a predictor $h(x_i, x_j, x_k)$, where $((x_t, y_t))_{i=1}^3$ is a random triplet of training examples.
  Given a test point $x^\dagger$, we form predictions by replicating and averaging, that is $f(x^\dagger) = \frac{1}{3} \sum_{t=1}^3 h(x^\dagger, x^\dagger, x^\dagger)_t$.
  %
  %
  \item \textbf{Radial Basis Function}, or RBF \citep{broomhead1988radial}, is a variant of ERM where we transform the logit vector $z \mapsto e^{-z^2}$ before passing them to the final softmax layer.
  In such a way, as the logit norm $\|z\| \to \infty$, the predicted softmax vector tends to the maximum entropy solution $(\frac{1}{C})_{c=1}^C$, signaling high uncertainty far away from the training data.
  RBFs have been proposed as defense to adversarial examples \citep{goodfellow2014explaining}, but they remain under-explored given the difficulties involved in their training.
  \item \textbf{Soft labeler} \citep{hinton2015distilling, szegedy2016rethinking} is a variant of ERM where the one-hot vector labels $y_i$ are \emph{smoothed} such that every zero becomes $\ell_\text{min} > 0$ and the solitary one becomes $\ell_\text{max} < 1$.
  Softening labels avoids saturation in the final softmax layer in neural network predictors, one of the main causes of overly-confident predictors.
  Using soft labels, we can identify softmax vectors with entries exceeding $\ell_\text{max}$ as ``over-shoots'', regarding them as uncertain extrapolations.
  %
  %
\end{itemize}

\paragraph{Ensembles of predictors} We also consider ensembles of predictors trained by each of the algorithms above.
Ensembles are commonly regarded as the state-of-the-art in uncertainty estimation \citep{lakshminarayanan2016simple}.
In particular, and for each algorithm, we construct bagging ensembles by (i) selecting the best $K \in \{1, 5\}$ predictors $\{f^k\}_{k=1}^K$ from all considered random initializations and hyper-parameters, and (ii) returning the average function $f(x^\dagger) := \frac{1}{K} \sum_{k=1}^M f^k(x^\dagger)$.

\section{Uncertainty measures}
\label{sec:measures}

Using \benchmark{}, we can equip a trained predictor $f$ with six different uncertainty measures.
An uncertainty measure is a real-valued function $u(f, x^\dagger)$ designed to return small values for \emph{in-domain} instances $x^\dagger$, and large values for \emph{out-domain} instances $x^\dagger$.
To describe the different measures, let $\{s_{(1)}, \ldots, s_{(C)}\}$ be the softmax scores returned by $f(x^\dagger)$ sorted in decreasing order.
\begin{itemize}
  \item \textbf{Largest} \citep{hendrycks2016baseline} returns minus the largest softmax score, $-s_{(1)}$ 
  \item \textbf{Softmax gap} \citep{tagasovska2018single} returns $s_{(2)} - s_{(1)}$.
  \item \textbf{Entropy} \citep{shannon2001mathematical} returns $- \sum_{c=1}^C s_{(c)} \mathbb{I}\{s_{(c)} > 0\} \log s_{(c)}$.
  \item \textbf{Norm of the Jacobian} \citep{novak2018sensitivity} returns $\|\nabla_x f(x^\dagger)\|^2_2$.
  \item \textbf{GMM} \citep{mukhoti2021deterministic} estimates one Gaussian density $\mathcal{N}(\phi(x); \mu_c, \Sigma_c)$ per-class, on top of the feature vectors $\phi(x)$ collected from a in-domain validation set.
  Given a test example $x^\dagger$, return $-\sum_{c=1}^C \lambda_c \cdot \mathcal{N}(\phi(x^\dagger); \mu_c, \Sigma_c)$, where $\lambda_c$ is the proportion of in-domain validation examples from class $c$.
  \item \textbf{Test-time augmentation} \citep{ashukha2020pitfalls} returns $-\max_c (\frac{1}{A} \sum_{a=1}^A f(x^\dagger_a))_c$.
  This is the measure ``\textbf{Largest}'' about the average prediction over $A$ random data augmentations $\{x^\dagger_a\}_{a=1}^A$ of the test instance $x^\dagger$.
\end{itemize}

These uncertainty measures are applicable to all the algorithms considered in Section~\ref{sec:algorithms}.
Additionally, some algorithms provide their \textbf{Native} uncertainty measures, outlined below.
\begin{itemize}
  \item For \textbf{Mixup}, we return
  $\| \lambda \cdot f(x^\dagger) + (1 - \lambda) \cdot \bar{y}  - f(\lambda \cdot x^\dagger + (1 - \lambda) \cdot \bar{x})\|_2^2$,
  where we recall that $\lambda \sim \text{Beta}(\alpha, \alpha)$, and $(\bar{x}, \bar{y})$ is the average image and label from the training set.
  This measures if the test example $x^\dagger$ violates the Mixup criterion wrt the training dataset average.
  \item For \textbf{RND} and \textbf{OC}, we return
  $\| w_{\text{student}}(\phi(x^\dagger)) - w_{\text{teacher}}(\phi(x^\dagger))\|_2^2$,
  that is, we consider a prediction uncertain if the outputs of the student and teacher disagree.
  We expect this disagreement to be related predictive uncertainty, as the student did not observe the behaviour of the teacher at out-domain instances $x^\dagger$.
  \item \textbf{For Soft labeler} we return $(s_{(1)} - \ell_\text{max})^2$.
  This measures the discrepancy between the largest softmax and the positive soft label target, able to signal overly-confident predictions.
  \item For \textbf{MC-Dropout} and \textbf{Ensembles}, and following~\citep{lakshminarayanan2016simple}, we return the Jensen-Shannon divergence between the $K$ members (or stochastic forward passes) $f^1, \ldots, f^K$ of the ensemble:
  \begin{equation*}
      H\left(\frac{1}{K} \sum_{k=1}^K f^k(x^\dagger)\right) - \frac{1}{K} \sum_{k=1}^K H(f^k(x^\dagger)).
  \end{equation*}
\end{itemize}

Note that the models ERM, OC, and RND are equivalent in all aspects except in comparisons involving the uncertainty measure Native.
This is because the only difference between these three models is the training of an external student for RND and OC, used only in their Native uncertainty measures.

\section{Evaluation metrics}
\label{sec:metrics}

For each algorithm-measure pair, we evaluate several metrics both \emph{in-domain} and \emph{out-domain}.

\subsection{In-domain metrics}
\label{sec:metrics-in}

Following~\citep{havasi2021training}, we implement four metrics to assess the performance and calibration of predictors when facing in-domain test examples.
\begin{itemize}
  \item \textbf{Top-1} and \textbf{Top-5} classification accuracy \citep{russakovsky2015imagenet}.
  \item \textbf{Expected Calibration Error} or ECE \citep{guo2017calibration}:
\begin{equation*}
  \frac{1}{B} \sum_{b=1}^B \frac{|B_b|}{n} \left| \text{acc}(f, B_b) - \text{conf}(f, B_b) \right|,
\end{equation*}
where $B_b$ contains the examples where the algorithm predicts a softmax score of $b$.
The functions \emph{acc} and \emph{conf} compute the average classification accuracy and largest softmax score of $f$ over $B_b$.
In a nutshell, ECE is minimized when $f$ is calibrated, that is, $f$ is wrong $p\%$ of the times it predicts a largest softmax score $p$.
Following \citep{guo2017calibration}, we discrete $b \in [0, 1]$ into $15$ equally-spaced bins.
\item \textbf{Negative Log Likelihood (NLL)} Also known as the cross-entropy loss, this is the objective minimized during the training process of the algorithms.
\end{itemize}

\subsection{Out-domain metrics}
\label{sec:metrics-out}

After measuring the performance of a predictor in-domain, we equip it with an uncertainty measure. 
We assess the uncertainty estimates of each predictor-measure pair using three metrics:
\begin{itemize}
  \item \textbf{Area Under the Curve}, or AUC \citep{tagasovska2018single}, describes how well does the predictor-measure pair distinguish between in-domain and out-domain examples over all thresholds of the uncertainty measure.
  \item \textbf{Confusion matrix at fixed threshold}. To reject out-domain examples in real scenarios, one must fix a threshold $\theta$ for the selected uncertainty measure.
  We do so by computing the 95\% quantile of the uncertainty measure, computed over an in-domain validation set.
  Then, at testing time, we declare one example out-domain if the uncertainty measure exceeds $\theta$.
  This strategy is equivalent to the statistical hypothesis test with null ``$H_0$: the observed example is in-domain''.
  To understand where does the uncertainty measure hit or miss, we monitor two metrics:
  \textbf{InAsIn} (percentage of in-domain examples classified as in-domain),
  \textbf{OutAsOut} (percentage of in-domain examples classified as out-domain).
  Please note that From these two metrics, we can deduce:
  \textbf{InAsOut} (percentage in-domain examples classified as out-domain, also known as false positives or type-I errors),
  \textbf{OutAsIn} (percentage out-domain examples classified as in-domain, also known as false negatives or type-II errors).
\end{itemize}

\section{Ablations}
\label{sec:ablations}

We execute our entire test-bed under three additional ablations often discussed in the literature of uncertainty estimation.
\begin{itemize}
  \item We study the effect of \textbf{calibration}  by temperature scaling \citep{platt1999probabilistic}.
  To this end, we introduce a temperature scaling $\tau > 0$ before the softmax layer, resulting in predictions $\text{Softmax}(\frac{z}{\tau})$ about the logit vector $z$.
  We estimate the optimal temperature $\hat{\tau}$ by minimizing the \textbf{NLL} of the predictor across an in-domain validation set.
  We evaluate all metrics for both the un-calibrated ($\tau = 1)$ and calibrated ($\tau = \hat{\tau}$) predictors.
  According to previous literature \citep{guo2017calibration}, calibrated models provide better in-domain uncertainty estimates.
  \item We analyze the impact of \textbf{spectral normalization} applied to the featurizer $\phi$.
  Several recent works \citep{liu2020simple, van2020uncertainty, van2021improving, mukhoti2021deterministic} have highlighted the importance of controlling both the \emph{smoothness} and \emph{sensitivity} of the feature extraction process to achieve high-quality uncertainty estimates.
    On the one hand, enforcing \emph{smoothness} \emph{upper} \emph{bounds} the Lipschitz constant of $\phi$, limiting its reaction to changes in the input. 
  Smoothness is often enforced by normalizing each weight matrix in $\phi$ by its spectral norm \citep{miyato2018spectral}.
  On the other hand, enforcing \emph{sensitivity} \emph{lower} \emph{bounds} the Lipschitz constant of $\phi$, ensuring that the feature space reacts in some amount when the input changes.
  Sensitivity is often enforced by residual connections~\citep{he2016deep}, present in the ResNet models that we will use throughout our experiments.
\item We analyze the impact of the model size, running experiments with both ResNet-18 and ResNet-50 models~\citep{he2016deep}.
\end{itemize}

\section{Experimental protocol}
\label{sec:experiments}

We are now ready to conduct experiments on the ImageNot benchmark for uncertainty estimation (Section~\ref{sec:datasets}) for all combinations of algorithms (Section~\ref{sec:algorithms}) and measures~(Section~\ref{sec:measures}), that we will evaluate under all metrics~(Section~\ref{sec:metrics}) and ablations (Section~\ref{sec:ablations}). 

\emph{Hyper-parameter search.}\quad We train each algorithm sixty times, arising from a combination of (i) ResNet-18 or ResNet-50 architectures (ii) using or not spectral normalization, (iii) five hyper-parameter trials, and (iv) three random train/validation splits of the in-domain data (data seeds).
We opt for a random hyper-parameter search~\citep{bergstra2012random}, where the search grid for each algorithm is detailed in Table~\ref{table:hyperparameters}.
More specifically, while the first trial uses the default hyper-parameter configuration suggested by the authors of each algorithm, the additional four trials explore random hyper-parameters.
\begin{table}
\begin{center}
\begin{tabular}{llll}
  \toprule
  \textbf{Algorithm}            & \textbf{Hyper-parameter}  & \textbf{Default value} & \textbf{Random search distribution}\\
  \midrule
  \multirow{3}{*}{all}          & learning rate             & 0.1                    & $10^{\text{Uniform}(-2, -0.3)}$\\
                                & momentum                  & 0.9                    & $\text{Choice}([0.5, 0.9, 0.99])$\\
                                & weight decay              & $10^{-4}$              & $10^{\text{Uniform}(-5, -3)}$\\
  \midrule
  \multirow{1}{*}{Mixup}        & mixing parameter          & 0.3                    & $\text{Choice}([0.1, 0.2, 0.3, 1, 2])$\\
  \midrule
  \multirow{2}{*}{MC-Dropout}   & dropout rate              & 0.05                   & $\text{Choice}([0.05, 0.1, 0.2])$\\
                                & number of passes          & 10                     & $\text{Choice}([10])$\\
  \midrule
  \multirow{3}{*}{MIMO}         & number of subnetworks     & 2                      & $\text{RandInt}(2, 5)$\\
                                & prob. input repetition    & 0.6                    & $\text{Uniform}(0, 1)$\\
                                & batch repetition          & 2                      & $\text{RandInt}(1, 5)$\\
  \midrule
  \multirow{3}{*}{RND, OC}      & teacher width             & 128                    & $\text{Choice}([64, 128, 256])$\\
                                & teacher depth             & 3                      & $\text{Choice}([2, 3, 4])$\\
                                & regularization            & 0                      & $10^{\text{Uniform}(-2, 1)}$\\
  \midrule
  \multirow{1}{*}{Soft labeler} & soft label value          & 128                    & $\text{Choice}([0.7, 0.8, 0.9])$\\
  \bottomrule
\end{tabular}
\end{center}
\caption{Default hyper-parameters and random search grids for all algorithms.}
\label{table:hyperparameters}
\end{table}

\emph{Model selection.}\quad After training all instances of a given algorithm, we report the average and standard deviation (over data seeds) for all metrics of two chosen models.
On the one hand, we report metrics for the \emph{best model} ($k=1$), which is the one minimizing the average (over data seeds) log-likelihood in the in-domain validation set. 
On the other hand, we report metrics also for the \emph{ensemble model} formed by the five different hyper-parameter trials.
Finally, we report all results of both the best and the ensemble model with and without calibration and for all model sizes~(Section~\ref{sec:ablations}).

\emph{Software and hardware.}\quad All algorithms are implemented in PyTorch~\citep{paszke2019pytorch} and train ResNet-18 or ResNet-50 backbones~\citep{he2016deep} initialized by following~\citep{glorot2010understanding}.
In addition, the Gaussian process in DUE is implemented in GPyTorch~\citep{gardner2018gpytorch}.
These models are trained by stochastic gradient descent~\citep{robbins1951stochastic, bottou2012stochastic} for 100 epochs, with mini-batches of 256 examples distributed over 8 NVIDIA Tesla V100 GPUs (yielding an effective mini-batch of size 32 per GPU).
During training, we decay the learning rate by a factor of 10 every 30 epochs.

\section{Conclusions}
\label{sec:conclusion}

Table~\ref{table:in-domain-ResNet50} summarizes our results for in-domain metrics (Section~\ref{sec:metrics-in}), and Table~\ref{table:out-domain-ResNet50-Entropy} summarizes our results for out-domain metrics (Section~\ref{sec:metrics-out}) and the best performing measure (Entropy, see Section~\ref{sec:measures}).
These tables contain the results for all algorithms~(Section~\ref{sec:algorithms}), ablations (use of spectral normalization, use of temperature calibration; see Section~\ref{sec:ablations}), and ensembling (whether bagging the best $k=1$ or $k=5$ models) for the largest model (ResNet50) considered.
Appendix~\ref{app:in-18} contains the in-domain resutls for ResNet18, and Appendix~\ref{app:all-measures} contains the out-domain result tables for all the other measures.
From the \textbf{in-domain} results summarized in Table~\ref{table:in-domain-ResNet50}, we identify the following key takeaways:
\begin{itemize}
  \item No algorithm significantly outperforms ERM in any in-domain metric.
  \item Ensembling multiple models ($k=5$) improves all in-domain metrics.
  \item Temperature calibration helps, decreasing the expected calibration error by up to $30\%$.
  \item Spectral normalization has a neutral and marginal effect on in-domain metrics.
  \item Most algorithms are able to achieve similar performances, as their hyper-parameter searches allow them to behave like ERM. 
\end{itemize}
Next, from the \textbf{out-domain} results summarized in Table~\ref{table:out-domain-ResNet50-Entropy}, we identify the following key takeaways:
\begin{itemize}
  \item A single MIMO is the only classifier outperforming ensembles of ERM classifiers on the OutAsOut metric, or the rate of correct identification of out-domain instances. 
  \item No other algorithm significantly outperforms ERM in any out-domain metric.
  \item Ensembling models ($k=5$) improves all out-domain metrics for all classifiers but MIMO.
  \item Temperature calibration has a neutral and marginal effect on out-domain metrics.
  \item Spectral normalization has a marginal negative effect on out-domain metrics.
  \item Most algorithms are able to achieve similar performances, as their hyper-parameter searches allow them to behave like ERM. 
\end{itemize}

Appendix~\ref{app:all-measures}, summarizing the \textbf{out-domain} performance of all other uncertainty measures, provides us with additional takeaways, in particular:
\begin{itemize}
  \item The best performing uncertainty measures, are Entropy, Largest, and Gap, with Entropy taking the lead with a generous gap.
  \item The uncertainty measures Augmentations and Jacobian exhibit a poor performance.
  \item Increasing model size leads to large improvements in the out-domain metric OutAsOut.
  \item Given enough capacity, a single MIMO classifier surpasses ensembles of 5 ERM classifiers on the out-domain metric OutAsOut.
  \item The Native uncertainty measures (those specific measures provided by some algorithms) do not exhibit a good performance.
\end{itemize}

From all of these results, our recommendation us to \emph{use calibrated ensembles of ERM classifiers} or \emph{a single MIMO classifier}, with no need for spectral normalization, and allowing the largest model size. When computational resources do not allow large model sizes or the use of ensembles, we recommend to use \emph{a single calibrated ERM models}.

\paragraph{Other results}
Despite our best efforts, we were unable to obtain competitive performances when using the algorithms RBF, \citep{broomhead1988radial}
or the uncertainty measure GMM \citep{mukhoti2021deterministic}.
We believe that training RBFs
at this large scale are challenging optimization problems that deserve further study in our community.
Furthermore, we believe that the large number of classes in our study (266 ImageNET classes instead of the 10 CIFAR-10 classes often considered) makes for a difficult problem for density-based uncertainty measures such as GMM.


\begin{table}[h!]
\begin{center}
\resizebox{1\textwidth}{!}{%
\begin{tabular}{llllllll}
\toprule
\textbf{algorithm}& \textbf{spectral}& \textbf{calibration}& \textbf{k}& \textbf{ACC@1}& \textbf{ACC@5}& \textbf{ECE}& \textbf{NLL}\\
\midrule
\multirow{8}{*}{ERM} & \multirow{4}{*}{False} & \multirow{2}{*}{initial} & 1.0 &  $0.762 \pm 0.004$ &  $0.859 \pm 0.001$ &  $0.050 \pm 0.003$ &  $0.952 \pm 0.013$ \\
            &       &         & 5.0 &  $0.782 \pm 0.001$ &  $0.871 \pm 0.002$ &  $0.030 \pm 0.001$ &  $0.855 \pm 0.001$ \\
            &       & \multirow{2}{*}{learned} & 1.0 &  $0.763 \pm 0.001$ &  $0.859 \pm 0.004$ &  $0.046 \pm 0.001$ &  $0.945 \pm 0.009$ \\
            &       &         & 5.0 &  $0.782 \pm 0.001$ &  $0.871 \pm 0.002$ &  $0.034 \pm 0.007$ &  $0.864 \pm 0.016$ \\
            & \multirow{4}{*}{True} & \multirow{2}{*}{initial} & 1.0 &  $0.764 \pm 0.002$ &  $0.859 \pm 0.001$ &  $0.054 \pm 0.002$ &  $0.957 \pm 0.009$ \\
            &       &         & 5.0 &  $0.782 \pm 0.001$ &  $0.872 \pm 0.002$ &  $0.031 \pm 0.001$ &  $0.853 \pm 0.003$ \\
            &       & \multirow{2}{*}{learned} & 1.0 &  $0.764 \pm 0.002$ &  $0.859 \pm 0.001$ &  $0.045 \pm 0.001$ &  $0.947 \pm 0.011$ \\
\bottomrule
\end{tabular}
}\end{center}
\caption{In-domain results for backbone ResNet50.}\label{table:in-domain-ResNet50}
\end{table}

\clearpage
\newpage

\begin{table}[h!]
\begin{center}
\resizebox{0.9\textwidth}{!}{%
\begin{tabular}{lllllll}
\toprule
\textbf{algorithm}& \textbf{spectral}& \textbf{calibration}& \textbf{k}& \textbf{AUC}& \textbf{InAsIn}& \textbf{OutAsOut}\\
\midrule
\multirow{8}{*}{ERM} & \multirow{4}{*}{False} & \multirow{2}{*}{initial} & 1.0 &  $0.869 \pm 0.003$ &  $0.941 \pm 0.001$ &  $0.382 \pm 0.012$ \\
            &       &         & 5.0 &  $0.874 \pm 0.001$ &  $0.941 \pm 0.002$ &  $0.375 \pm 0.008$ \\
            &       & \multirow{2}{*}{learned} & 1.0 &  $0.864 \pm 0.004$ &  $0.942 \pm 0.001$ &  $0.359 \pm 0.009$ \\
            &       &         & 5.0 &  $0.864 \pm 0.000$ &  $0.940 \pm 0.002$ &  $0.366 \pm 0.006$ \\
            & \multirow{4}{*}{True} & \multirow{2}{*}{initial} & 1.0 &  $0.858 \pm 0.007$ &  $0.943 \pm 0.001$ &  $0.354 \pm 0.022$ \\
            &       &         & 5.0 &  $0.872 \pm 0.002$ &  $0.941 \pm 0.002$ &  $0.371 \pm 0.007$ \\
            &       & \multirow{2}{*}{learned} & 1.0 &  $0.860 \pm 0.008$ &  $0.943 \pm 0.001$ &  $0.358 \pm 0.023$ \\
\bottomrule
\end{tabular}
}\end{center}
\caption{Out-domain results for measure Entropy and backbone ResNet50.}\label{table:out-domain-ResNet50-Entropy}
\end{table}

\clearpage
\newpage

\bibliography{uncertanity_nets}

\begin{thebibliography}{50}
\providecommand{\natexlab}[1]{#1}
\providecommand{\url}[1]{\texttt{#1}}
\expandafter\ifx\csname urlstyle\endcsname\relax
  \providecommand{\doi}[1]{doi: #1}\else
  \providecommand{\doi}{doi: \begingroup \urlstyle{rm}\Url}\fi

\bibitem[Abdar et~al.(2020)Abdar, Pourpanah, Hussain, Rezazadegan, Liu,
  Ghavamzadeh, Fieguth, Khosravi, Acharya, Makarenkov, et~al.]{abdar2020review}
Moloud Abdar, Farhad Pourpanah, Sadiq Hussain, Dana Rezazadegan, Li~Liu,
  Mohammad Ghavamzadeh, Paul Fieguth, Abbas Khosravi, U~Rajendra Acharya,
  Vladimir Makarenkov, et~al.
\newblock A review of uncertainty quantification in deep learning: Techniques,
  applications and challenges.
\newblock \emph{arXiv}, 2020.

\bibitem[Alvarez-Melis and Jaakkola(2017)]{alvarez2017causal}
David Alvarez-Melis and Tommi~S Jaakkola.
\newblock A causal framework for explaining the predictions of black-box
  sequence-to-sequence models.
\newblock \emph{arXiv}, 2017.

\bibitem[Ashukha et~al.(2020)Ashukha, Lyzhov, Molchanov, and
  Vetrov]{ashukha2020pitfalls}
Arsenii Ashukha, Alexander Lyzhov, Dmitry Molchanov, and Dmitry Vetrov.
\newblock Pitfalls of in-domain uncertainty estimation and ensembling in deep
  learning.
\newblock \emph{arXiv}, 2020.

\bibitem[Begoli et~al.(2019)Begoli, Bhattacharya, and Kusnezov]{begoli2019need}
Edmon Begoli, Tanmoy Bhattacharya, and Dimitri Kusnezov.
\newblock The need for uncertainty quantification in machine-assisted medical
  decision making.
\newblock \emph{Nature Machine Intelligence}, 2019.

\bibitem[Bergstra and Bengio(2012)]{bergstra2012random}
James Bergstra and Yoshua Bengio.
\newblock Random search for hyper-parameter optimization.
\newblock \emph{Journal of machine learning research}, 13\penalty0 (2), 2012.

\bibitem[Bottou(2012)]{bottou2012stochastic}
L{\'e}on Bottou.
\newblock Stochastic gradient descent tricks.
\newblock In \emph{Neural networks: Tricks of the trade}, pages 421--436.
  Springer, 2012.

\bibitem[Broomhead and Lowe(1988)]{broomhead1988radial}
David~S Broomhead and David Lowe.
\newblock Radial basis functions, multi-variable functional interpolation and
  adaptive networks.
\newblock Technical report, Royal Signals and Radar Establishment Malvern
  (United Kingdom), 1988.

\bibitem[Burda et~al.(2018)Burda, Edwards, Storkey, and
  Klimov]{burda2018exploration}
Yuri Burda, Harrison Edwards, Amos Storkey, and Oleg Klimov.
\newblock Exploration by random network distillation.
\newblock \emph{arXiv}, 2018.

\bibitem[Chalapathy and Chawla(2019)]{chalapathy2019deep}
Raghavendra Chalapathy and Sanjay Chawla.
\newblock Deep learning for anomaly detection: A survey.
\newblock \emph{arXiv}, 2019.

\bibitem[Gal and Ghahramani(2016)]{gal2016dropout}
Yarin Gal and Zoubin Ghahramani.
\newblock Dropout as a bayesian approximation: Representing model uncertainty
  in deep learning.
\newblock In \emph{ICML}, 2016.

\bibitem[Gardner et~al.(2018)Gardner, Pleiss, Bindel, Weinberger, and
  Wilson]{gardner2018gpytorch}
Jacob~R Gardner, Geoff Pleiss, David Bindel, Kilian~Q Weinberger, and
  Andrew~Gordon Wilson.
\newblock Gpytorch: Blackbox matrix-matrix gaussian process inference with gpu
  acceleration.
\newblock \emph{arXiv preprint arXiv:1809.11165}, 2018.

\bibitem[Glorot and Bengio(2010)]{glorot2010understanding}
Xavier Glorot and Yoshua Bengio.
\newblock Understanding the difficulty of training deep feedforward neural
  networks.
\newblock In \emph{Proceedings of the thirteenth international conference on
  artificial intelligence and statistics}, pages 249--256. JMLR Workshop and
  Conference Proceedings, 2010.

\bibitem[Goodfellow et~al.(2014)Goodfellow, Shlens, and
  Szegedy]{goodfellow2014explaining}
Ian~J Goodfellow, Jonathon Shlens, and Christian Szegedy.
\newblock Explaining and harnessing adversarial examples.
\newblock \emph{arXiv}, 2014.

\bibitem[Gulrajani and Lopez-Paz(2020)]{gulrajani2020search}
Ishaan Gulrajani and David Lopez-Paz.
\newblock In search of lost domain generalization.
\newblock \emph{arXiv}, 2020.

\bibitem[Guo et~al.(2017)Guo, Pleiss, Sun, and Weinberger]{guo2017calibration}
Chuan Guo, Geoff Pleiss, Yu~Sun, and Kilian~Q Weinberger.
\newblock On calibration of modern neural networks.
\newblock In \emph{ICML}, 2017.

\bibitem[Havasi et~al.(2021)Havasi, Jenatton, Fort, Liu, Snoek,
  Lakshminarayanan, Dai, and Tran]{havasi2021training}
Marton Havasi, Rodolphe Jenatton, Stanislav Fort, Jeremiah~Zhe Liu, Jasper
  Snoek, Balaji Lakshminarayanan, Andrew~Mingbo Dai, and Dustin Tran.
\newblock Training independent subnetworks for robust prediction.
\newblock In \emph{ICLR}, 2021.

\bibitem[He et~al.(2016)He, Zhang, Ren, and Sun]{he2016deep}
Kaiming He, Xiangyu Zhang, Shaoqing Ren, and Jian Sun.
\newblock Deep residual learning for image recognition.
\newblock In \emph{CVPR}, 2016.

\bibitem[Hein et~al.(2019)Hein, Andriushchenko, and Bitterwolf]{hein2019relu}
Matthias Hein, Maksym Andriushchenko, and Julian Bitterwolf.
\newblock Why relu networks yield high-confidence predictions far away from the
  training data and how to mitigate the problem.
\newblock In \emph{CVPR}, 2019.

\bibitem[Henaff et~al.(2019)Henaff, Canziani, and LeCun]{henaff2019model}
Mikael Henaff, Alfredo Canziani, and Yann LeCun.
\newblock Model-predictive policy learning with uncertainty regularization for
  driving in dense traffic.
\newblock \emph{arXiv}, 2019.

\bibitem[Hendrycks and Dietterich(2019)]{hendrycks2019benchmarking}
Dan Hendrycks and Thomas Dietterich.
\newblock Benchmarking neural network robustness to common corruptions and
  perturbations.
\newblock \emph{arXiv}, 2019.

\bibitem[Hendrycks and Gimpel(2016)]{hendrycks2016baseline}
Dan Hendrycks and Kevin Gimpel.
\newblock A baseline for detecting misclassified and out-of-distribution
  examples in neural networks.
\newblock \emph{arXiv}, 2016.

\bibitem[Hendrycks et~al.(2021)Hendrycks, Zhao, Basart, Steinhardt, and
  Song]{hendrycks2021natural}
Dan Hendrycks, Kevin Zhao, Steven Basart, Jacob Steinhardt, and Dawn Song.
\newblock Natural adversarial examples.
\newblock In \emph{Proceedings of the IEEE/CVF Conference on Computer Vision
  and Pattern Recognition}, pages 15262--15271, 2021.

\bibitem[Hinton et~al.(2015)Hinton, Vinyals, and Dean]{hinton2015distilling}
Geoffrey Hinton, Oriol Vinyals, and Jeff Dean.
\newblock Distilling the knowledge in a neural network.
\newblock \emph{arXiv}, 2015.

\bibitem[Lakshminarayanan et~al.(2016)Lakshminarayanan, Pritzel, and
  Blundell]{lakshminarayanan2016simple}
Balaji Lakshminarayanan, Alexander Pritzel, and Charles Blundell.
\newblock Simple and scalable predictive uncertainty estimation using deep
  ensembles.
\newblock \emph{NeurIPS}, 2016.

\bibitem[LeCun et~al.(2015)LeCun, Bengio, and Hinton]{lecun2015deep}
Yann LeCun, Yoshua Bengio, and Geoffrey Hinton.
\newblock Deep learning.
\newblock \emph{nature}, 2015.

\bibitem[Liu et~al.(2020)Liu, Lin, Padhy, Tran, Bedrax-Weiss, and
  Lakshminarayanan]{liu2020simple}
Jeremiah~Zhe Liu, Zi~Lin, Shreyas Padhy, Dustin Tran, Tania Bedrax-Weiss, and
  Balaji Lakshminarayanan.
\newblock Simple and principled uncertainty estimation with deterministic deep
  learning via distance awareness.
\newblock \emph{arXiv}, 2020.

\bibitem[Michelmore et~al.(2018)Michelmore, Kwiatkowska, and
  Gal]{michelmore2018evaluating}
Rhiannon Michelmore, Marta Kwiatkowska, and Yarin Gal.
\newblock Evaluating uncertainty quantification in end-to-end autonomous
  driving control.
\newblock \emph{arXiv}, 2018.

\bibitem[Miyato et~al.(2018)Miyato, Kataoka, Koyama, and
  Yoshida]{miyato2018spectral}
Takeru Miyato, Toshiki Kataoka, Masanori Koyama, and Yuichi Yoshida.
\newblock Spectral normalization for generative adversarial networks.
\newblock \emph{arXiv}, 2018.

\bibitem[Mukhoti et~al.(2021)Mukhoti, Kirsch, van Amersfoort, Torr, and
  Gal]{mukhoti2021deterministic}
Jishnu Mukhoti, Andreas Kirsch, Joost van Amersfoort, Philip~HS Torr, and Yarin
  Gal.
\newblock Deterministic neural networks with appropriate inductive biases
  capture epistemic and aleatoric uncertainty.
\newblock \emph{arXiv}, 2021.

\bibitem[Novak et~al.(2018)Novak, Bahri, Abolafia, Pennington, and
  Sohl-Dickstein]{novak2018sensitivity}
Roman Novak, Yasaman Bahri, Daniel~A Abolafia, Jeffrey Pennington, and Jascha
  Sohl-Dickstein.
\newblock Sensitivity and generalization in neural networks: an empirical
  study.
\newblock \emph{arXiv}, 2018.

\bibitem[Ovadia et~al.(2019)Ovadia, Fertig, Ren, Nado, Sculley, Nowozin,
  Dillon, Lakshminarayanan, and Snoek]{ovadia2019can}
Yaniv Ovadia, Emily Fertig, Jie Ren, Zachary Nado, David Sculley, Sebastian
  Nowozin, Joshua~V Dillon, Balaji Lakshminarayanan, and Jasper Snoek.
\newblock Can you trust your model's uncertainty? evaluating predictive
  uncertainty under dataset shift.
\newblock \emph{arXiv}, 2019.

\bibitem[Paszke et~al.(2019)Paszke, Gross, Massa, Lerer, Bradbury, Chanan,
  Killeen, Lin, Gimelshein, Antiga, et~al.]{paszke2019pytorch}
Adam Paszke, Sam Gross, Francisco Massa, Adam Lerer, James Bradbury, Gregory
  Chanan, Trevor Killeen, Zeming Lin, Natalia Gimelshein, Luca Antiga, et~al.
\newblock Pytorch: An imperative style, high-performance deep learning library.
\newblock \emph{arXiv}, 2019.

\bibitem[Platt et~al.(1999)]{platt1999probabilistic}
John Platt et~al.
\newblock Probabilistic outputs for support vector machines and comparisons to
  regularized likelihood methods.
\newblock \emph{Advances in Large Margin Classifiers}, 1999.

\bibitem[Robbins and Monro(1951)]{robbins1951stochastic}
Herbert Robbins and Sutton Monro.
\newblock A stochastic approximation method.
\newblock \emph{The annals of mathematical statistics}, pages 400--407, 1951.

\bibitem[Ruff et~al.(2021)Ruff, Kauffmann, Vandermeulen, Montavon, Samek,
  Kloft, Dietterich, and M{\"u}ller]{ruff2021unifying}
Lukas Ruff, Jacob~R Kauffmann, Robert~A Vandermeulen, Gr{\'e}goire Montavon,
  Wojciech Samek, Marius Kloft, Thomas~G Dietterich, and Klaus-Robert
  M{\"u}ller.
\newblock A unifying review of deep and shallow anomaly detection.
\newblock \emph{Proceedings of the IEEE}, 2021.

\bibitem[Russakovsky et~al.(2015)Russakovsky, Deng, Su, Krause, Satheesh, Ma,
  Huang, Karpathy, Khosla, Bernstein, et~al.]{russakovsky2015imagenet}
Olga Russakovsky, Jia Deng, Hao Su, Jonathan Krause, Sanjeev Satheesh, Sean Ma,
  Zhiheng Huang, Andrej Karpathy, Aditya Khosla, Michael Bernstein, et~al.
\newblock Imagenet large scale visual recognition challenge.
\newblock \emph{IJCV}, 2015.

\bibitem[Santurkar et~al.(2020)Santurkar, Tsipras, and
  Madry]{santurkar2020breeds}
Shibani Santurkar, Dimitris Tsipras, and Aleksander Madry.
\newblock Breeds: Benchmarks for subpopulation shift.
\newblock \emph{arXiv}, 2020.

\bibitem[Settles(2009)]{settles2009active}
Burr Settles.
\newblock Active learning literature survey.
\newblock 2009.

\bibitem[Shannon(1948)]{shannon2001mathematical}
Claude~Elwood Shannon.
\newblock A mathematical theory of communication.
\newblock \emph{Bell System Technical Journal}, 1948.

\bibitem[Srivastava et~al.(2014)Srivastava, Hinton, Krizhevsky, Sutskever, and
  Salakhutdinov]{srivastava2014dropout}
Nitish Srivastava, Geoffrey Hinton, Alex Krizhevsky, Ilya Sutskever, and Ruslan
  Salakhutdinov.
\newblock Dropout: a simple way to prevent neural networks from overfitting.
\newblock \emph{The journal of machine learning research}, 2014.

\bibitem[Szegedy et~al.(2016)Szegedy, Vanhoucke, Ioffe, Shlens, and
  Wojna]{szegedy2016rethinking}
Christian Szegedy, Vincent Vanhoucke, Sergey Ioffe, Jon Shlens, and Zbigniew
  Wojna.
\newblock Rethinking the inception architecture for computer vision.
\newblock In \emph{CVPR}, 2016.

\bibitem[Tagasovska and Lopez-Paz(2018)]{tagasovska2018single}
Natasa Tagasovska and David Lopez-Paz.
\newblock Single-model uncertainties for deep learning.
\newblock \emph{arXiv}, 2018.

\bibitem[Thulasidasan et~al.(2019)Thulasidasan, Chennupati, Bilmes,
  Bhattacharya, and Michalak]{thulasidasan2019mixup}
Sunil Thulasidasan, Gopinath Chennupati, Jeff Bilmes, Tanmoy Bhattacharya, and
  Sarah Michalak.
\newblock On mixup training: Improved calibration and predictive uncertainty
  for deep neural networks.
\newblock \emph{arXiv}, 2019.

\bibitem[Ulmer and Cin{\`a}(2020)]{ulmer2020know}
Dennis Ulmer and Giovanni Cin{\`a}.
\newblock Know your limits: Monotonicity \& softmax make neural classifiers
  overconfident on ood data.
\newblock \emph{arXiv}, 2020.

\bibitem[Van~Amersfoort et~al.(2020)Van~Amersfoort, Smith, Teh, and
  Gal]{van2020uncertainty}
Joost Van~Amersfoort, Lewis Smith, Yee~Whye Teh, and Yarin Gal.
\newblock Uncertainty estimation using a single deep deterministic neural
  network.
\newblock In \emph{ICML}, 2020.

\bibitem[van Amersfoort et~al.(2021)van Amersfoort, Smith, Jesson, Key, and
  Gal]{van2021improving}
Joost van Amersfoort, Lewis Smith, Andrew Jesson, Oscar Key, and Yarin Gal.
\newblock Improving deterministic uncertainty estimation in deep learning for
  classification and regression.
\newblock \emph{arXiv}, 2021.

\bibitem[Vapnik(1992)]{vapnik1992principles}
Vladimir Vapnik.
\newblock Principles of risk minimization for learning theory.
\newblock In \emph{NeurIPS}, 1992.

\bibitem[Wadoux(2019)]{wadoux2019using}
Alexandre MJ-C Wadoux.
\newblock Using deep learning for multivariate mapping of soil with quantified
  uncertainty.
\newblock \emph{Geoderma}, 2019.

\bibitem[Ward~Jr(1963)]{ward1963hierarchical}
Joe~H Ward~Jr.
\newblock Hierarchical grouping to optimize an objective function.
\newblock \emph{Journal of the American statistical association}, 1963.

\bibitem[Zhang et~al.(2017)Zhang, Cisse, Dauphin, and
  Lopez-Paz]{zhang2017mixup}
Hongyi Zhang, Moustapha Cisse, Yann~N Dauphin, and David Lopez-Paz.
\newblock mixup: Beyond empirical risk minimization.
\newblock \emph{arXiv}, 2017.

\end{thebibliography}
\bibliographystyle{plainnat}

\clearpage
\newpage
\appendix

\section{In-domain results}
\label{app:in-18}
\begin{table}[h!]
\begin{center}
\resizebox{0.9\textwidth}{!}{%
\begin{tabular}{llllllll}
\toprule
\textbf{algorithm}& \textbf{spectral}& \textbf{calibration}& \textbf{k}& \textbf{ACC@1}& \textbf{ACC@5}& \textbf{ECE}& \textbf{NLL}\\
\midrule
\multirow{8}{*}{ERM} & \multirow{4}{*}{False} & \multirow{2}{*}{initial} & 1.0 &  $0.718 \pm 0.001$ &  $0.821 \pm 0.001$ &  $0.044 \pm 0.001$ &  $1.139 \pm 0.002$ \\
            &       &         & 5.0 &  $0.745 \pm 0.001$ &  $0.843 \pm 0.002$ &  $0.044 \pm 0.001$ &  $1.017 \pm 0.001$ \\
            &       & \multirow{2}{*}{learned} & 1.0 &  $0.718 \pm 0.001$ &  $0.821 \pm 0.001$ &  $0.050 \pm 0.002$ &  $1.145 \pm 0.003$ \\
            &       &         & 5.0 &  $0.745 \pm 0.001$ &  $0.843 \pm 0.002$ &  $0.049 \pm 0.003$ &  $1.048 \pm 0.004$ \\
            & \multirow{4}{*}{True} & \multirow{2}{*}{initial} & 1.0 &  $0.720 \pm 0.001$ &  $0.822 \pm 0.001$ &  $0.046 \pm 0.004$ &  $1.144 \pm 0.004$ \\
            &       &         & 5.0 &  $0.744 \pm 0.001$ &  $0.843 \pm 0.001$ &  $0.045 \pm 0.000$ &  $1.020 \pm 0.003$ \\
            &       & \multirow{2}{*}{learned} & 1.0 &  $0.720 \pm 0.001$ &  $0.822 \pm 0.001$ &  $0.052 \pm 0.003$ &  $1.151 \pm 0.003$ \\
\bottomrule
\end{tabular}
}\end{center}
\caption{In-domain results for backbone ResNet18.}\label{table:in-domain-ResNet18}
\end{table}

\clearpage
\newpage

\section{Out-domain results}
\label{app:all-measures}
\begin{table}[h!]
\begin{center}
\resizebox{0.85\textwidth}{!}{%
\begin{tabular}{lllllll}
\toprule
\textbf{algorithm}& \textbf{spectral}& \textbf{calibration}& \textbf{k}& \textbf{AUC}& \textbf{InAsIn}& \textbf{OutAsOut}\\
\midrule
\multirow{8}{*}{ERM} & \multirow{4}{*}{False} & \multirow{2}{*}{initial} & 1.0 &  $0.644 \pm 0.017$ &  $0.949 \pm 0.001$ &  $0.093 \pm 0.016$ \\
            &       &         & 5.0 &  $0.668 \pm 0.011$ &  $0.966 \pm 0.030$ &  $0.076 \pm 0.066$ \\
            &       & \multirow{2}{*}{learned} & 1.0 &  $0.649 \pm 0.017$ &  $0.952 \pm 0.001$ &  $0.090 \pm 0.017$ \\
            &       &         & 5.0 &  $0.658 \pm 0.009$ &  $0.949 \pm 0.001$ &  $0.105 \pm 0.011$ \\
            & \multirow{4}{*}{True} & \multirow{2}{*}{initial} & 1.0 &  $0.648 \pm 0.010$ &  $0.949 \pm 0.000$ &  $0.097 \pm 0.004$ \\
            &       &         & 5.0 &  $0.664 \pm 0.005$ &  $0.948 \pm 0.003$ &  $0.105 \pm 0.006$ \\
            &       & \multirow{2}{*}{learned} & 1.0 &  $0.649 \pm 0.010$ &  $0.950 \pm 0.001$ &  $0.097 \pm 0.003$ \\
\bottomrule
\end{tabular}
}\end{center}
\caption{Out-domain results for measure Augmentations and backbone ResNet18.}\label{table:out-domain-ResNet18-Augmentations}
\end{table}

\begin{table}[h!]
\begin{center}
\resizebox{0.9\textwidth}{!}{%
\begin{tabular}{lllllll}
\toprule
\textbf{algorithm}& \textbf{spectral}& \textbf{calibration}& \textbf{k}& \textbf{AUC}& \textbf{InAsIn}& \textbf{OutAsOut}\\
\midrule
\multirow{8}{*}{ERM} & \multirow{4}{*}{False} & \multirow{2}{*}{initial} & 1.0 &  $0.663 \pm 0.004$ &  $0.951 \pm 0.001$ &  $0.106 \pm 0.010$ \\
            &       &         & 5.0 &  $0.682 \pm 0.004$ &  $0.966 \pm 0.030$ &  $0.079 \pm 0.068$ \\
            &       & \multirow{2}{*}{learned} & 1.0 &  $0.657 \pm 0.010$ &  $0.951 \pm 0.002$ &  $0.105 \pm 0.017$ \\
            &       &         & 5.0 &  $0.662 \pm 0.004$ &  $0.948 \pm 0.002$ &  $0.106 \pm 0.003$ \\
            & \multirow{4}{*}{True} & \multirow{2}{*}{initial} & 1.0 &  $0.655 \pm 0.002$ &  $0.951 \pm 0.001$ &  $0.105 \pm 0.006$ \\
            &       &         & 5.0 &  $0.662 \pm 0.007$ &  $0.948 \pm 0.002$ &  $0.101 \pm 0.011$ \\
            &       & \multirow{2}{*}{learned} & 1.0 &  $0.657 \pm 0.001$ &  $0.947 \pm 0.003$ &  $0.113 \pm 0.009$ \\
\bottomrule
\end{tabular}
}\end{center}
\caption{Out-domain results for measure Augmentations and backbone ResNet50.}\label{table:out-domain-ResNet50-Augmentations}
\end{table}

\begin{table}[h!]
\begin{center}
\resizebox{0.9\textwidth}{!}{%
\begin{tabular}{lllllll}
\toprule
\textbf{algorithm}& \textbf{spectral}& \textbf{calibration}& \textbf{k}& \textbf{AUC}& \textbf{InAsIn}& \textbf{OutAsOut}\\
\midrule
\multirow{8}{*}{ERM} & \multirow{4}{*}{False} & \multirow{2}{*}{initial} & 1.0 &  $0.811 \pm 0.009$ &  $0.941 \pm 0.001$ &  $0.258 \pm 0.011$ \\
            &       &         & 5.0 &  $0.831 \pm 0.007$ &  $0.939 \pm 0.002$ &  $0.273 \pm 0.012$ \\
            &       & \multirow{2}{*}{learned} & 1.0 &  $0.810 \pm 0.009$ &  $0.941 \pm 0.001$ &  $0.257 \pm 0.011$ \\
            &       &         & 5.0 &  $0.822 \pm 0.006$ &  $0.937 \pm 0.002$ &  $0.274 \pm 0.012$ \\
            & \multirow{4}{*}{True} & \multirow{2}{*}{initial} & 1.0 &  $0.809 \pm 0.006$ &  $0.941 \pm 0.001$ &  $0.244 \pm 0.008$ \\
            &       &         & 5.0 &  $0.830 \pm 0.006$ &  $0.938 \pm 0.001$ &  $0.267 \pm 0.013$ \\
            &       & \multirow{2}{*}{learned} & 1.0 &  $0.808 \pm 0.005$ &  $0.941 \pm 0.001$ &  $0.245 \pm 0.009$ \\
\bottomrule
\end{tabular}
}\end{center}
\caption{Out-domain results for measure Entropy and backbone ResNet18.}\label{table:out-domain-ResNet18-Entropy}
\end{table}

\begin{table}[h!]
\begin{center}
\resizebox{0.9\textwidth}{!}{%
\begin{tabular}{lllllll}
\toprule
\textbf{algorithm}& \textbf{spectral}& \textbf{calibration}& \textbf{k}& \textbf{AUC}& \textbf{InAsIn}& \textbf{OutAsOut}\\
\midrule
\multirow{8}{*}{ERM} & \multirow{4}{*}{False} & \multirow{2}{*}{initial} & 1.0 &  $0.775 \pm 0.007$ &  $0.944 \pm 0.002$ &  $0.177 \pm 0.010$ \\
            &       &         & 5.0 &  $0.795 \pm 0.004$ &  $0.942 \pm 0.001$ &  $0.192 \pm 0.008$ \\
            &       & \multirow{2}{*}{learned} & 1.0 &  $0.774 \pm 0.007$ &  $0.944 \pm 0.002$ &  $0.176 \pm 0.010$ \\
            &       &         & 5.0 &  $0.785 \pm 0.004$ &  $0.942 \pm 0.002$ &  $0.178 \pm 0.008$ \\
            & \multirow{4}{*}{True} & \multirow{2}{*}{initial} & 1.0 &  $0.778 \pm 0.003$ &  $0.944 \pm 0.003$ &  $0.172 \pm 0.011$ \\
            &       &         & 5.0 &  $0.794 \pm 0.004$ &  $0.939 \pm 0.001$ &  $0.186 \pm 0.009$ \\
            &       & \multirow{2}{*}{learned} & 1.0 &  $0.778 \pm 0.002$ &  $0.944 \pm 0.003$ &  $0.171 \pm 0.011$ \\
\bottomrule
\end{tabular}
}\end{center}
\caption{Out-domain results for measure Gap and backbone ResNet18.}\label{table:out-domain-ResNet18-Gap}
\end{table}

\begin{table}[h!]
\begin{center}
\resizebox{0.9\textwidth}{!}{%
\begin{tabular}{lllllll}
\toprule
\textbf{algorithm}& \textbf{spectral}& \textbf{calibration}& \textbf{k}& \textbf{AUC}& \textbf{InAsIn}& \textbf{OutAsOut}\\
\midrule
\multirow{8}{*}{ERM} & \multirow{4}{*}{False} & \multirow{2}{*}{initial} & 1.0 &  $0.822 \pm 0.001$ &  $0.945 \pm 0.002$ &  $0.223 \pm 0.000$ \\
            &       &         & 5.0 &  $0.829 \pm 0.002$ &  $0.944 \pm 0.001$ &  $0.225 \pm 0.005$ \\
            &       & \multirow{2}{*}{learned} & 1.0 &  $0.819 \pm 0.003$ &  $0.945 \pm 0.001$ &  $0.216 \pm 0.006$ \\
            &       &         & 5.0 &  $0.820 \pm 0.003$ &  $0.944 \pm 0.002$ &  $0.209 \pm 0.007$ \\
            & \multirow{4}{*}{True} & \multirow{2}{*}{initial} & 1.0 &  $0.815 \pm 0.004$ &  $0.944 \pm 0.001$ &  $0.214 \pm 0.005$ \\
            &       &         & 5.0 &  $0.827 \pm 0.002$ &  $0.942 \pm 0.003$ &  $0.223 \pm 0.005$ \\
            &       & \multirow{2}{*}{learned} & 1.0 &  $0.817 \pm 0.005$ &  $0.944 \pm 0.001$ &  $0.218 \pm 0.006$ \\
\bottomrule
\end{tabular}
}\end{center}
\caption{Out-domain results for measure Gap and backbone ResNet50.}\label{table:out-domain-ResNet50-Gap}
\end{table}

\begin{table}[h!]
\begin{center}
\resizebox{0.9\textwidth}{!}{%
\begin{tabular}{lllllll}
\toprule
\textbf{algorithm}& \textbf{spectral}& \textbf{calibration}& \textbf{k}& \textbf{AUC}& \textbf{InAsIn}& \textbf{OutAsOut}\\
\midrule
\multirow{8}{*}{ERM} & \multirow{4}{*}{False} & \multirow{2}{*}{initial} & 1.0 &  $0.552 \pm 0.007$ &  $0.956 \pm 0.001$ &  $0.045 \pm 0.002$ \\
            &       &         & 5.0 &  $0.218 \pm 0.012$ &  $0.960 \pm 0.000$ &  $0.001 \pm 0.000$ \\
            &       & \multirow{2}{*}{learned} & 1.0 &  $0.552 \pm 0.007$ &  $0.956 \pm 0.001$ &  $0.045 \pm 0.002$ \\
            &       &         & 5.0 &  $0.218 \pm 0.012$ &  $0.960 \pm 0.000$ &  $0.001 \pm 0.000$ \\
            & \multirow{4}{*}{True} & \multirow{2}{*}{initial} & 1.0 &  $0.566 \pm 0.029$ &  $0.957 \pm 0.001$ &  $0.051 \pm 0.009$ \\
            &       &         & 5.0 &  $0.219 \pm 0.009$ &  $0.959 \pm 0.001$ &  $0.001 \pm 0.000$ \\
            &       & \multirow{2}{*}{learned} & 1.0 &  $0.566 \pm 0.029$ &  $0.957 \pm 0.001$ &  $0.051 \pm 0.009$ \\
\bottomrule
\end{tabular}
}\end{center}
\caption{Out-domain results for measure Jacobian and backbone ResNet18.}\label{table:out-domain-ResNet18-Jacobian}
\end{table}

\begin{table}[h!]
\begin{center}
\resizebox{0.9\textwidth}{!}{%
\begin{tabular}{lllllll}
\toprule
\textbf{algorithm}& \textbf{spectral}& \textbf{calibration}& \textbf{k}& \textbf{AUC}& \textbf{InAsIn}& \textbf{OutAsOut}\\
\midrule
\multirow{4}{*}{ERM} & \multirow{2}{*}{False} & initial & 1.0 &  $0.545 \pm 0.036$ &  $0.951 \pm 0.003$ &  $0.058 \pm 0.013$ \\
            &       & learned & 1.0 &  $0.534 \pm 0.029$ &  $0.951 \pm 0.001$ &  $0.059 \pm 0.012$ \\
            & \multirow{2}{*}{True} & initial & 1.0 &  $0.522 \pm 0.020$ &  $0.950 \pm 0.003$ &  $0.053 \pm 0.004$ \\
            &       & learned & 1.0 &  $0.522 \pm 0.020$ &  $0.950 \pm 0.003$ &  $0.053 \pm 0.004$ \\
\multirow{4}{*}{MIMO} & \multirow{2}{*}{False} & initial & 1.0 &  $0.192 \pm 0.007$ &  $0.955 \pm 0.002$ &  $0.003 \pm 0.001$ \\
            &       & learned & 1.0 &  $0.192 \pm 0.007$ &  $0.955 \pm 0.002$ &  $0.003 \pm 0.001$ \\
            & \multirow{2}{*}{True} & initial & 1.0 &  $0.194 \pm 0.009$ &  $0.955 \pm 0.001$ &  $0.003 \pm 0.000$ \\
\bottomrule
\end{tabular}
}\end{center}
\caption{Out-domain results for measure Jacobian and backbone ResNet50.}\label{table:out-domain-ResNet50-Jacobian}
\end{table}

\begin{table}[h!]
\begin{center}
\resizebox{0.9\textwidth}{!}{%
\begin{tabular}{lllllll}
\toprule
\textbf{algorithm}& \textbf{spectral}& \textbf{calibration}& \textbf{k}& \textbf{AUC}& \textbf{InAsIn}& \textbf{OutAsOut}\\
\midrule
\multirow{8}{*}{ERM} & \multirow{4}{*}{False} & \multirow{2}{*}{initial} & 1.0 &  $0.795 \pm 0.009$ &  $0.940 \pm 0.002$ &  $0.245 \pm 0.006$ \\
            &       &         & 5.0 &  $0.819 \pm 0.005$ &  $0.937 \pm 0.002$ &  $0.271 \pm 0.009$ \\
            &       & \multirow{2}{*}{learned} & 1.0 &  $0.794 \pm 0.009$ &  $0.940 \pm 0.002$ &  $0.244 \pm 0.006$ \\
            &       &         & 5.0 &  $0.804 \pm 0.005$ &  $0.937 \pm 0.000$ &  $0.253 \pm 0.005$ \\
            & \multirow{4}{*}{True} & \multirow{2}{*}{initial} & 1.0 &  $0.796 \pm 0.004$ &  $0.941 \pm 0.000$ &  $0.235 \pm 0.007$ \\
            &       &         & 5.0 &  $0.818 \pm 0.005$ &  $0.937 \pm 0.001$ &  $0.263 \pm 0.011$ \\
            &       & \multirow{2}{*}{learned} & 1.0 &  $0.795 \pm 0.004$ &  $0.941 \pm 0.000$ &  $0.234 \pm 0.007$ \\
\bottomrule
\end{tabular}
}\end{center}
\caption{Out-domain results for measure Largest and backbone ResNet18.}\label{table:out-domain-ResNet18-Largest}
\end{table}

\begin{table}[h!]
\begin{center}
\resizebox{0.9\textwidth}{!}{%
\begin{tabular}{lllllll}
\toprule
\textbf{algorithm}& \textbf{spectral}& \textbf{calibration}& \textbf{k}& \textbf{AUC}& \textbf{InAsIn}& \textbf{OutAsOut}\\
\midrule
\multirow{8}{*}{ERM} & \multirow{4}{*}{False} & \multirow{2}{*}{initial} & 1.0 &  $0.846 \pm 0.002$ &  $0.942 \pm 0.001$ &  $0.336 \pm 0.010$ \\
            &       &         & 5.0 &  $0.856 \pm 0.001$ &  $0.940 \pm 0.000$ &  $0.342 \pm 0.003$ \\
            &       & \multirow{2}{*}{learned} & 1.0 &  $0.843 \pm 0.003$ &  $0.943 \pm 0.003$ &  $0.323 \pm 0.004$ \\
            &       &         & 5.0 &  $0.841 \pm 0.002$ &  $0.941 \pm 0.001$ &  $0.311 \pm 0.006$ \\
            & \multirow{4}{*}{True} & \multirow{2}{*}{initial} & 1.0 &  $0.837 \pm 0.006$ &  $0.942 \pm 0.000$ &  $0.313 \pm 0.014$ \\
            &       &         & 5.0 &  $0.854 \pm 0.001$ &  $0.941 \pm 0.003$ &  $0.334 \pm 0.005$ \\
            &       & \multirow{2}{*}{learned} & 1.0 &  $0.840 \pm 0.007$ &  $0.943 \pm 0.001$ &  $0.317 \pm 0.014$ \\
\bottomrule
\end{tabular}
}\end{center}
\caption{Out-domain results for measure Largest and backbone ResNet50.}\label{table:out-domain-ResNet50-Largest}
\end{table}

\begin{table}[h!]
\begin{center}
\resizebox{0.9\textwidth}{!}{%
\begin{tabular}{lllllll}
\toprule
\textbf{algorithm}& \textbf{spectral}& \textbf{calibration}& \textbf{k}& \textbf{AUC}& \textbf{InAsIn}& \textbf{OutAsOut}\\
\midrule
\multirow{4}{*}{ERM} & \multirow{2}{*}{False} & initial & 5.0 &  $0.768 \pm 0.004$ &  $0.942 \pm 0.002$ &  $0.126 \pm 0.002$ \\
            &       & learned & 5.0 &  $0.768 \pm 0.004$ &  $0.942 \pm 0.002$ &  $0.126 \pm 0.002$ \\
            & \multirow{2}{*}{True} & initial & 5.0 &  $0.771 \pm 0.007$ &  $0.943 \pm 0.003$ &  $0.128 \pm 0.009$ \\
            &       & learned & 5.0 &  $0.771 \pm 0.007$ &  $0.943 \pm 0.003$ &  $0.128 \pm 0.009$ \\
\multirow{8}{*}{MCDropout} & \multirow{4}{*}{False} & \multirow{2}{*}{initial} & 1.0 &  $0.519 \pm 0.000$ &  $0.960 \pm 0.001$ &  $0.094 \pm 0.003$ \\
            &       &         & 5.0 &  $0.772 \pm 0.003$ &  $0.945 \pm 0.002$ &  $0.126 \pm 0.002$ \\
            &       & \multirow{2}{*}{learned} & 1.0 &  $0.517 \pm 0.004$ &  $0.958 \pm 0.003$ &  $0.099 \pm 0.004$ \\
\bottomrule
\end{tabular}
}\end{center}
\caption{Out-domain results for measure Native and backbone ResNet18.}\label{table:out-domain-ResNet18-Native}
\end{table}

\begin{table}[h!]
\begin{center}
\resizebox{0.9\textwidth}{!}{%
\begin{tabular}{lllllll}
\toprule
\textbf{algorithm}& \textbf{spectral}& \textbf{calibration}& \textbf{k}& \textbf{AUC}& \textbf{InAsIn}& \textbf{OutAsOut}\\
\midrule
\multirow{4}{*}{ERM} & \multirow{2}{*}{False} & initial & 5.0 &  $0.792 \pm 0.005$ &  $0.941 \pm 0.001$ &  $0.144 \pm 0.003$ \\
            &       & learned & 5.0 &  $0.792 \pm 0.005$ &  $0.941 \pm 0.001$ &  $0.144 \pm 0.003$ \\
            & \multirow{2}{*}{True} & initial & 5.0 &  $0.793 \pm 0.002$ &  $0.942 \pm 0.000$ &  $0.143 \pm 0.006$ \\
            &       & learned & 5.0 &  $0.793 \pm 0.002$ &  $0.942 \pm 0.000$ &  $0.143 \pm 0.006$ \\
\multirow{8}{*}{MCDropout} & \multirow{4}{*}{False} & \multirow{2}{*}{initial} & 1.0 &  $0.523 \pm 0.003$ &  $0.968 \pm 0.001$ &  $0.094 \pm 0.004$ \\
            &       &         & 5.0 &  $0.792 \pm 0.005$ &  $0.942 \pm 0.003$ &  $0.148 \pm 0.009$ \\
            &       & \multirow{2}{*}{learned} & 1.0 &  $0.523 \pm 0.003$ &  $0.968 \pm 0.001$ &  $0.094 \pm 0.004$ \\
\bottomrule
\end{tabular}
}\end{center}
\caption{Out-domain results for measure Native and backbone ResNet50.}\label{table:out-domain-ResNet50-Native}
\end{table}

\clearpage
\newpage

\section{ImageNot dataset}
\label{app:partitions}

Our ImageNot dataset is a class partition of the ImageNet ILSVRC2012 dataset \citep{russakovsky2015imagenet}.

\subsection{In-domain classes}

\begin{multicols}{6}
\begin{verbatim}
n02666196
n02669723
n02672831
n02690373
n02699494
n02776631
n02783161
n02786058
n02791124
n02793495
n02794156
n02797295
n02799071
n02804610
n02808304
n02808440
n02814860
n02817516
n02834397
n02835271
n02837789
n02840245
n02859443
n02860847
n02869837
n02871525
n02877765
n02883205
n02892767
n02894605
n02895154
n02909870
n02927161
n02951358
n02951585
n02966687
n02974003
n02977058
n02979186
n02980441
n02981792
n02992211
n03000684
n03026506
n03028079
n03032252
n03042490
n03045698
n03063599
n03075370
n03085013
n03095699
n03124170
n03127747
n03127925
n03131574
n03160309
n03180011
n03187595
n03201208
n03207743
n03207941
n03216828
n03223299
n03240683
n03249569
n03272010
n03272562
n03290653
n03291819
n03325584
n03337140
n03344393
n03347037
n03372029
n03384352
n03388549
n03393912
n03394916
n03417042
n03425413
n03445777
n03445924
n03447721
n03452741
n03459775
n03461385
n03467068
n03478589
n03481172
n03482405
n03495258
n03496892
n03498962
n03527444
n03534580
n03535780
n03584254
n03590841
n03594734
n03594945
n03627232
n03633091
n03657121
n03661043
n03670208
n03680355
n03690938
n03706229
n03709823
n03710193
n03710721
n03717622
n03720891
n03721384
n03729826
n03733281
n03733805
n03742115
n03743016
n03759954
n03761084
n03763968
n03769881
n03773504
n03775071
n03775546
n03777568
n03781244
n03782006
n03785016
n03786901
n03787032
n03788365
n03791053
n03793489
n03794056
n03803284
n03814639
n03841143
n03843555
n03857828
n03866082
n03868863
n03874293
n03874599
n03877845
n03884397
n03887697
n03891332
n03895866
n03903868
n03920288
n03924679
n03930313
n03933933
n03938244
n03947888
n03950228
n03956157
n03958227
n03967562
n03976657
n03982430
n03983396
n03995372
n04004767
n04008634
n04019541
n04023962
n04033995
n04037443
n04039381
n04041544
n04065272
n04067472
n04070727
n04081281
n04086273
n04090263
n04118776
n04131690
n04141327
n04146614
n04153751
n04154565
n04162706
n04179913
n04192698
n04201297
n04208210
n04209133
n04228054
n04235860
n04238763
n04243546
n04254120
n04254777
n04258138
n04259630
n04264628
n04266014
n04270147
n04273569
n04286575
n04311004
n04311174
n04317175
n04325704
n04326547
n04330267
n04332243
n04346328
n04347754
n04355933
n04356056
n04357314
n04371430
n04371774
n04372370
n04376876
n04399382
n04404412
n04409515
n04418357
n04447861
n04456115
n04458633
n04462240
n04465501
n04476259
n04482393
n04485082
n04487394
n04501370
n04517823
n04522168
n04525038
n04525305
n04540053
n04548280
n04548362
n04550184
n04552348
n04553703
n04560804
n04579145
n04584207
n04590129
n04591157
n04591713
n04592741
n04604644
n04612504
n04613696
n06359193
n07802026
n07930864
n09193705
n09246464
n09288635
n09332890
n09421951
n09472597
n10148035
n15075141
\end{verbatim}
\end{multicols}

\clearpage
\newpage
\subsection{Out-domain classes}

\begin{multicols}{6}
\begin{verbatim}
n01440764
n01443537
n01484850
n01491361
n01494475
n01496331
n01498041
n01514668
n01514859
n01518878
n01530575
n01531178
n01532829
n01534433
n01537544
n01558993
n01560419
n01580077
n01582220
n01592084
n01601694
n01608432
n01614925
n01616318
n01622779
n01629819
n01630670
n01631663
n01632458
n01632777
n01641577
n01644373
n01644900
n01664065
n01665541
n01667114
n01667778
n01669191
n01675722
n01677366
n01682714
n01685808
n01687978
n01688243
n01689811
n01692333
n01693334
n01694178
n01695060
n01697457
n01698640
n01704323
n01728572
n01728920
n01729322
n01729977
n01734418
n01735189
n01737021
n01739381
n01740131
n01742172
n01744401
n01748264
n01749939
n01751748
n01753488
n01755581
n01756291
n01768244
n01770081
n01770393
n01773157
n01773549
n01773797
n01774384
n01774750
n01775062
n01776313
n01784675
n01795545
n01796340
n01797886
n01798484
n01806143
n01806567
n01807496
n01817953
n01818515
n01819313
n01820546
n01824575
n01828970
n01829413
n01833805
n01843065
n01843383
n01847000
n01855032
n01855672
n01860187
n01871265
n01872401
n01873310
n01877812
n01882714
n01883070
n01910747
n01914609
n01917289
n01924916
n01930112
n01943899
n01944390
n01945685
n01950731
n01955084
n01968897
n01978287
n01978455
n01980166
n01981276
n01983481
n01984695
n01985128
n01986214
n01990800
n02002556
n02002724
n02006656
n02007558
n02009229
n02009912
n02011460
n02012849
n02013706
n02017213
n02018207
n02018795
n02025239
n02027492
n02028035
n02033041
n02037110
n02051845
n02056570
n02058221
n02066245
n02071294
n02074367
n02077923
n02085620
n02085782
n02085936
n02086079
n02086240
n02086646
n02086910
n02087046
n02087394
n02088094
n02088238
n02088364
n02088466
n02088632
n02089078
n02089867
n02089973
n02090379
n02090622
n02090721
n02091032
n02091134
n02091244
n02091467
n02091635
n02091831
n02092002
n02092339
n02093256
n02093428
n02093647
n02093754
n02093859
n02093991
n02094114
n02094258
n02094433
n02095314
n02095570
n02095889
n02096051
n02096177
n02096294
n02096437
n02096585
n02097047
n02097130
n02097209
n02097298
n02097474
n02097658
n02098105
n02098286
n02098413
n02099267
n02099429
n02099601
n02099712
n02099849
n02100236
n02100583
n02100735
n02100877
n02101006
n02101388
n02101556
n02102040
n02102177
n02102318
n02102480
n02102973
n02104029
n02104365
n02105056
n02105162
n02105251
n02105412
n02105505
n02105641
n02105855
n02106030
n02106166
n02106382
n02106550
n02106662
n02107142
n02107312
n02107574
n02107683
n02107908
n02108000
n02108089
n02108422
n02108551
n02108915
n02109047
n02109525
n02109961
n02110063
n02110185
n02110341
n02110627
n02110806
n02110958
n02111129
n02111277
n02111500
n02111889
n02112018
n02112137
n02112350
n02112706
n02113023
n02113186
n02113624
\end{verbatim}
\end{multicols}

\end{document}